\documentclass[11pt,a4paper]{article}
\usepackage[dvipdfmx]{graphicx}
\usepackage[hyperref]{emnlp2018}
\usepackage{times}
\usepackage{latexsym}
\usepackage{amsmath}
\usepackage{amssymb}
\usepackage{enumerate}
\usepackage{booktabs}
\usepackage{multirow}
\usepackage{xcolor,colortbl}
\usepackage{pifont}
\newcommand{\cmark}{\ding{51}}%

\usepackage{url}

\definecolor{my-gray}{RGB}{230,230,230}
\newcommand*{\belowrulesepcolor}[1]{%
  \noalign{%
    \kern-\belowrulesep
    \begingroup
      \color{#1}%
      \hrule height\belowrulesep
    \endgroup
  }%
}
\newcommand*{\aboverulesepcolor}[1]{%
  \noalign{%
    \begingroup
      \color{#1}%
      \hrule height\aboverulesep
    \endgroup
    \kern-\aboverulesep
  }%
}

\aclfinalcopy 

\title{Discriminative Learning of Open-Vocabulary Object Retrieval \\ and Localization by Negative Phrase Augmentation}
\author{Ryota Hinami$^{1, 2}$ and Shin'ichi Satoh$^{2, 1}$\\
$^{1}$The University of Tokyo, $^{2}$National Institute of Infomatics \\
{\tt\small hinami@nii.ac.jp, satoh@nii.ac.jp}
}

\date{}

\begin{document}
\maketitle
\begin{abstract} 
Thanks to the success of object detection technology, we can retrieve
objects of the specified classes even from huge image collections.
However, the current state-of-the-art object detectors (such as Faster
R-CNN) can only handle pre-specified classes.  
In addition, large amounts of
positive and negative visual samples are required for training.  In
this paper, we address the problem of open-vocabulary object retrieval
and localization, where the target object is specified by a textual query
(e.g., a word or phrase).  We first propose Query-Adaptive R-CNN, a
simple extension of Faster R-CNN adapted to open-vocabulary queries,
by transforming the text embedding vector into an object classifier and
localization regressor.  Then, for discriminative training, we then
propose negative phrase augmentation (NPA) to mine hard negative
samples which are visually similar to the query and at the same time
semantically mutually exclusive of the query.  The proposed method can
retrieve and localize objects specified by a textual query from one
million images in only 0.5 seconds with high precision.

\end{abstract} 

\section{Introduction} 
\label{sec:intro}
Our goal is to retrieve objects from large-scale image database 
and localize their spatial locations given a textual query.
The task of object retrieval and localization has many applications
such as spatial position-aware image searches~\cite{Hinami2017} and 
it recently has gathered much attention from researchers.
While much of the previous work mainly focused on object instance retrieval wherein the query is an 
image~\cite{Shen2012,Taoa,Tolias2015},
recent approaches~\cite{Aytar2014,Hinami2016} enable retrieval of
more generic concepts such as an object category.
Although such approaches are built on the recent successes of object detection including that of R-CNN~\cite{Girshick2014}, 
object detection methods can generally handle only 
closed sets of categories (e.g., PASCAL 20 classes),
which severely limits the variety of queries when they are used as retrieval systems.
Open-vocabulary object localization is also a hot topic and 
many approaches are proposed to solve this problem~\cite{Plummer2015,Chen2017}.
However, most of them are not scalable to make them useful 
for large-scale retrieval.


\begin{figure}[t] 
\vspace{0mm}
\begin{center}
\includegraphics[width=1.00\linewidth]{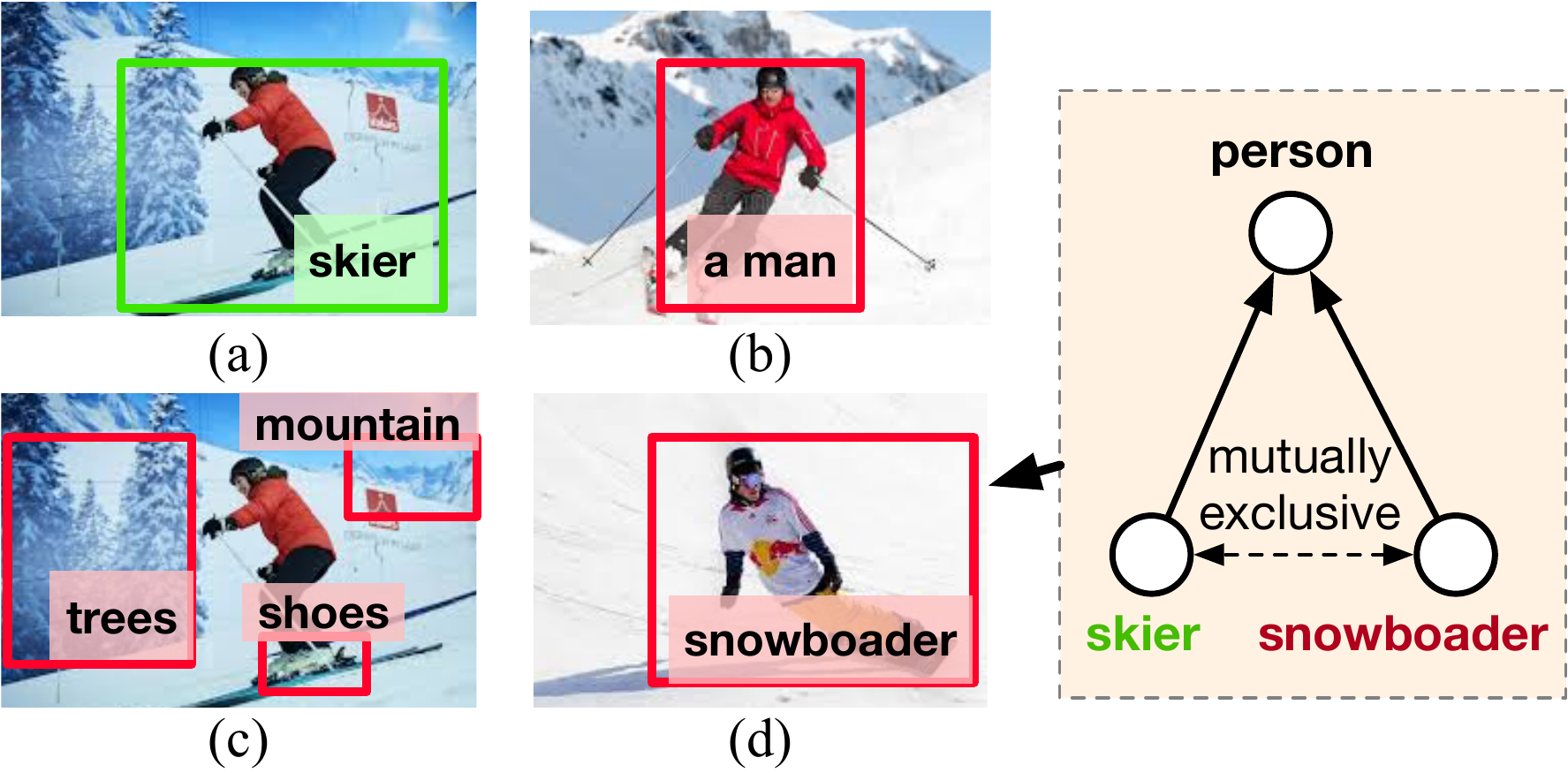}
\caption{Training examples in open-vocabulary object detection. 
(a) positive example of skier classifier. (b) examples without positive annotation, which can be positive.
(c) examples without positive annotation from an image that contains 
a positive example.
(d) proposed approach to select hard and true negative examples by using linguistics knowledge.
}
\label{fig:top} 
\vspace{0mm}
\end{center} 
\end{figure}

We first describe {\it Query-Adaptive} R-CNN 
as an extension of the Faster R-CNN~\cite{Ren2015} object detection 
framework to open-vocabulary object detection
simply by adding a component called a {\it detector generator}.
While Faster R-CNN learns the class-specific linear classifier as learnable
parameters of the neural network, we generate the weight of the classifier adaptively
from text descriptions by learning the detector generator 
(Fig.~\ref{fig:pipeline}b).
All of its components can be trained in an end-to-end manner.
In spite of its simple architecture, 
it outperforms all state-of-the-art methods  
in the Flickr30k Entities phrase localization task.
It can also be used for large-scale retrievals in the manner 
presented in \cite{Hinami2016}.

However, training a discriminative classifier is harder in the 
open-vocabulary setting.
Closed-vocabulary object detection models such as Faster R-CNN are trained using many negative examples, 
where a sufficient amount of good-quality negative examples is
shown to be important for learning a discriminative 
classifier~\cite{Felzenszwalb2010,Shrivastava2016}.
While closed-vocabulary object detection can use all regions without positive labels as negative data,
in open-vocabulary detection, it is not guaranteed that a region without a positive 
label is negative. 
For example, as shown in Fig.~\ref{fig:top}b,
a region with the annotation \texttt{a man} is not always negative for \texttt{skier}.
Since training data for open-vocabulary object detection is generally 
composed of images, each having region annotations with free 
descriptions,
it is nearly impossible to do an exhaustive annotation throughout 
the dataset for all possible descriptions.
Another possible approach is to use the regions without positive labels 
in the image that contains positive examples,
as shown in Fig.~\ref{fig:top}c.
Although they can be guaranteed to be positive by carefully annotating 
the datasets,
negative examples are only limited to the objects that cooccur with the learned class.


To exploit negative data in open-vocabulary object detection, 
we use mutually exclusive relationships between categories.
For example, an object with a label \texttt{dog} is guaranteed to be negative 
for the \texttt{cat} class because \texttt{dog} and \texttt{cat} are mutually exclusive.
In addition, we propose an approach to select {\it hard negative} 
phrases that are difficult to discriminate
(e.g., selecting \texttt{zebra} for \texttt{horse}).
This approach, called {\it negative phrase augmentation (NPA)},
significantly improves the discriminative ability of the classifier 
and improves the retrieval performance by a large margin.

Our contributions are as follows.
1) We propose Query-Adaptive R-CNN, an extension of Faster R-CNN 
to open vocabulary, 
that is a simple yet strong method of open-vocabulary object detection and
that outperforms all state-of-the-art methods in the phrase localization task.
2) We propose negative phrase augmentation (NPA) to exploit hard negative examples  
when training for open-vocabulary object detection,
which makes the classifier more discriminative and 
robust to distractors in retrieval.
Our method can accurately find objects amidst one million images in 0.5 second.


\section{Related work} 
{\bf Phrase localization.}
Object grounding with natural language descriptions has recently drawn much attention and 
several tasks and approaches have been proposed for it~\cite{Guadarrama2014,Hu2015f,Kazemzadeh2014,Mao2016,Plummer2015}.
The most related task to ours is the phrase localization introduced 
by Plummer et al.~\cite{Plummer2015}, whose goal is to localize objects
that corresponds to noun phrases in textual descriptions from an image.
Chen et al.~\cite{Chen2017} is the closest to our work in terms of  
learning region proposals and performing regression conditioned upon a query.
However, most phrase localization methods are not scalable and
cannot be used for retrieval tasks.
Some approaches~\cite{Plummer2017,Wang} learn a common 
subspace between the text and image for phrase localization.
Instead of learning the subspace between the image and sentence as in standard 
cross-modal searches, they learn the subspace between a region and a phrase.
In particular, Wang et al.~\cite{Wang} use a deep neural network to learn the joint embedding of images and text;
their training uses structure-preserving constraints based on structured matching.
Although these approaches can be used for large-scale retrieval,
their accuracy is not as good as recent state-of-the-art methods.

{\bf Object retrieval and localization.}
Object retrieval and localization have been researched
in the context of particular object retrieval~\cite{Shen2012,Taoa,Tolias2015}, 
where a query is given as an image.
Aytar et al.~\cite{Aytar2014} proposed retrieval and localization of 
generic category objects by extending the object detection technique to large-scale retrieval.
Hinami and Satoh~\cite{Hinami2016} extended the R-CNN to
large-scale retrieval by using approximate nearest neighbor search techniques.
However, they assumed that the detector of the category is given as a query and
require many sample images with bounding box annotations in order to learn the detector.
Several other approaches have used the external search engines (e.g., Google image search) 
to get training images from textual queries~\cite{Arandjelovi2012,Chatfield2015}.
Instead, we generate an object detector directly from the given textual query by using a neural network.

{\bf Parameter prediction by neural network.}
Query-Adaptive R-CNN generates the weights of the detector
from the query instead of learning them by backpropagation.
The dynamic filter network~\cite{DeBrabandere2016} is one of the first methods that generate
neural network parameters dynamically conditioned on an input.
Several subsequent approaches use this idea in zero-shot learning~\cite{Ba2016} 
and visual question answering~\cite{Noh2016}.
Zhang et al.~\cite{Zhang2017} integrates this idea into the Fast R-CNN
framework by dynamically generating the classifier from the text in a similar manner to \cite{Ba2016}.
We extend this work to the case of large-scale retrieval.
The proposed Query-Adaptive R-CNN generates the regressor weights 
and learn the region proposal network following Faster R-CNN.
It enables precise localization with fewer proposals, 
which makes the retrieval system more memory efficient.
In addition, we propose a novel hard negative mining approach,
called negative phrase augmentation, 
which makes the generated classifier more discriminative.


\section{Query-Adaptive R-CNN} 
\begin{figure}[t] 
\begin{center}
\includegraphics[width=1.00\linewidth]{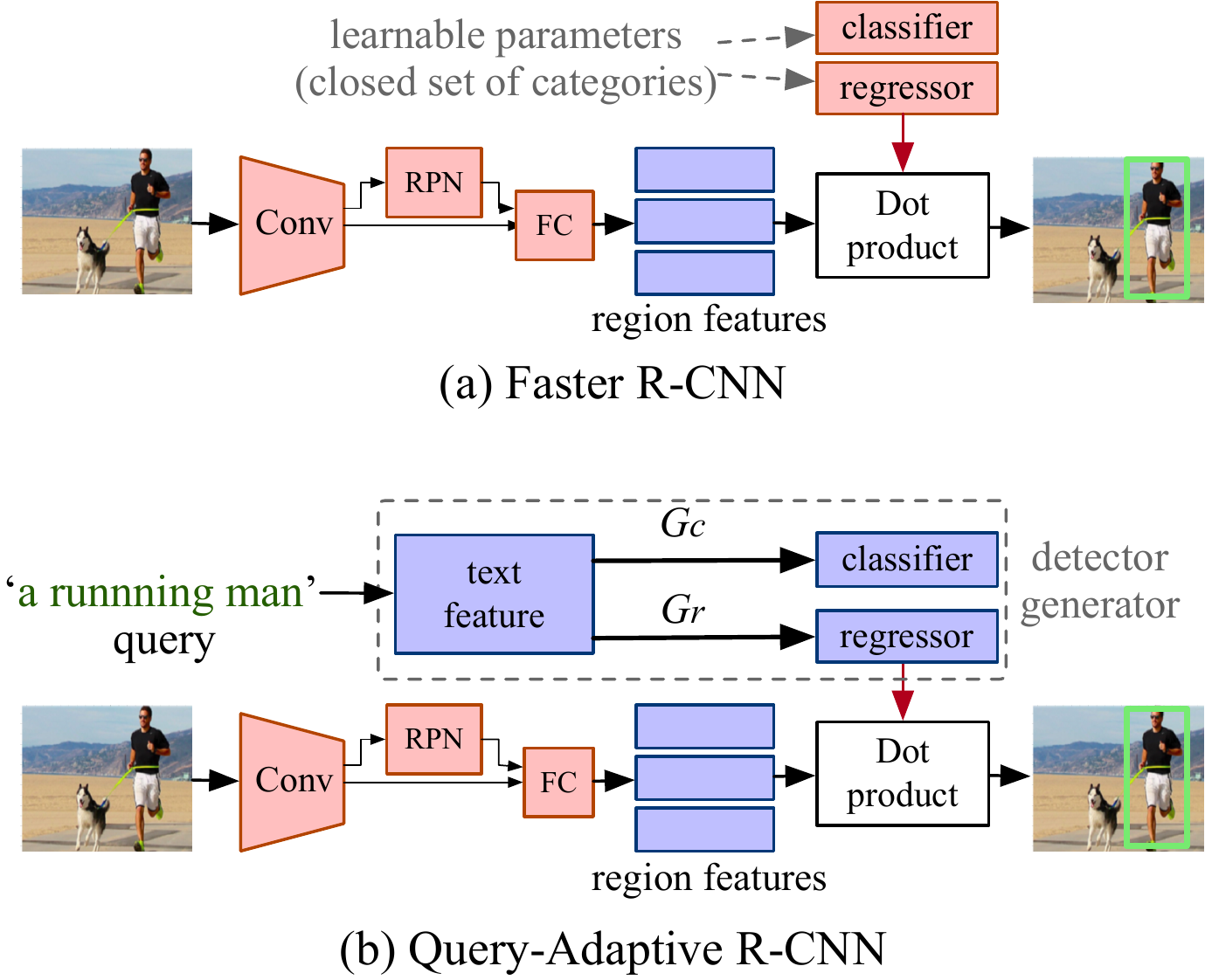}
\vspace{-5mm}
\caption{Difference in network architecture between (a) Faster R-CNN and (b) Query-Adaptive R-CNN.
While Faster R-CNN learns the classifier of a closed set of categories as learnable parameters of neural networks,
Query-Adaptive R-CNN generates a classifier and regressor adaptively from a query text by learning 
a detector generator that transforms the text into a classifier and regressor.
}
\vspace{-2mm}
\label{fig:pipeline} 
\end{center} 
\end{figure}

Query-adaptive R-CNN is a simple extension of Faster R-CNN 
to open-vocabulary object detection.
While Faster R-CNN detects objects of fixed categories,
Query-Adaptive R-CNN detects any objects specified by a textual
phrase. 
Figure~\ref{fig:pipeline} illustrates the difference between 
Faster R-CNN and Query-Adaptive R-CNN. 
While Faster R-CNN learns a class-specific classifier and regressor 
as parameters of the neural networks, 
Query-Adaptive R-CNN generates them from the query text by using a detector generator.
Query-Adaptive R-CNN is a simple but effective method that surpasses 
state-of-the-art phrase localization methods and can be easily extended 
to the case of large-scale retrieval.
Furthermore, its retrieval accuracy is significantly improved by a
novel training strategy called negative phrase augmentation (Sec.~\ref{sec:train}).


\subsection{Architecture}
\label{sec:qrcn_arch}
The network is composed of two subnetworks: a {\it region feature extractor}
and {\it detector generator}, both of which are trained in an end-to-end manner.
The region feature extractor takes an image as input and outputs 
features extracted from sub-regions that are candidate objects. 
Following Faster R-CNN~\cite{Ren2015}, regions are detected using a 
region proposal network (RPN) 
and the features of the last layer (e.g., fc7 in VGG network) 
are used as region features.
The detector generator takes a text description as an input 
and outputs a linear classifier and regressor for the description
(e.g., if \texttt{a dog} is given, \texttt{a dog} classifier and regressor are output). 
Finally, a confidence and a regressed bounding box are predicted for each region
by applying the classifier and regressor to the region features.


{\bf Detector generator.}
The detector generator transforms the given text $t$ into a classifier
$\mathbf{w}^c$ and regressor 
$(\mathbf{w}^r_x, \mathbf{w}^r_y, \mathbf{w}^r_w, \mathbf{w}^r_h)$,
where $\mathbf{w}_c$ is the weight of a linear classifier and $(\mathbf{w}^r_x, \mathbf{w}^r_y, \mathbf{w}^r_w, \mathbf{w}^r_h)$ 
is the weight of a linear regressor in terms of $x$, $y$, width $w$, and height $h$, following \cite{Girshick2014}.
We first transform a text $t$ of variable length into 
a text embedding vector $\mathbf{v}$.
Other phrase localization approaches uses 
the Fisher vector encoding of word2vec~\cite{Klein2015,Plummer2015} or 
long-short term memory (LSTM)~\cite{Chen2017}
for the phrase embedding. However, 
we found that the simple mean pooling of word2vec~\cite{Mikolov2013} 
performs better than these methods for our model
(comparisons given in the supplemental material).
The text embedding is then transformed into a detector, i.e.,
$\mathbf{w}_c=G_c(\mathbf{v})$ and $(\mathbf{w}^r_x, \mathbf{w}^r_y, \mathbf{w}^r_w, \mathbf{w}^r_h)=G_r(\mathbf{v})$.
Here, we use a linear transformation for $G_c$ 
(i.e., $\mathbf{w}_c=\mathbf{W}\mathbf{v}$, where $\mathbf{W}$ is a projection matrix). 
For the regressor,
we use a multi-layer perceptron with one hidden layer to predict each of 
$(\mathbf{w}^r_x, \mathbf{w}^r_y, \mathbf{w}^r_w, \mathbf{w}^r_h)=G_r(\mathbf{v})$.
We tested various architectures for $G_r$ and found that sharing the hidden layer and reducing the dimension 
of the hidden layer (up to $16$) does not adversely affect the performance,
while at the same time it significantly reduces the number of parameters (see Sec.~\ref{sec:exp_ph} for details).


\subsection{Training with Negative Phrase Augmentation}
\label{sec:train}
\begin{figure}[t] 
\begin{center}
\includegraphics[width=1.03\linewidth]{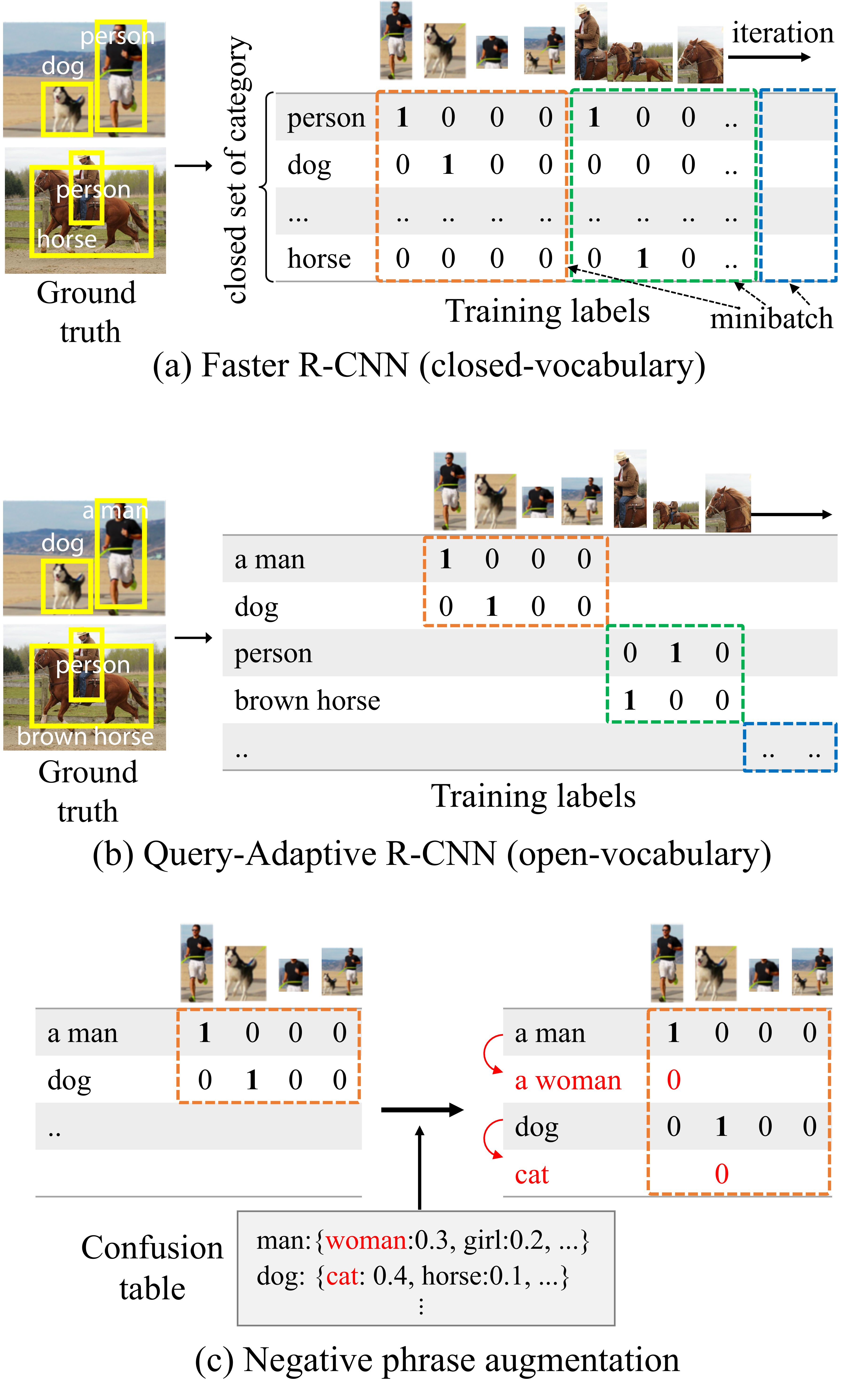}
\caption{Difference in training between (a) closed-vocabulary and (b) open-vocabulary object detection.
The approach of NPA is illustrated in (c).}
\label{fig:neg_aug} 
\end{center} 
\end{figure}

All components of Query-Adaptive R-CNN can be jointly trained 
in an end-to-end manner.
The training strategy basically follows that of Faster R-CNN. 
The differences are shown in Figure~\ref{fig:neg_aug}.
Faster R-CNN is trained with the fixed closed set of categories (Fig.~\ref{fig:neg_aug}a), 
where all regions without a positive label can be used as negative examples.
On the other hand,
Query-Adaptive R-CNN is trained using the open-vocabulary phrases annotated to the regions
(Fig.~\ref{fig:neg_aug}b), where sufficient negative examples cannot be used for each phrase 
compared to Faster R-CNN
because a region without a positive label is not guaranteed to be negative
in open-vocabulary object detection.
We solve this problem by proposing negative phrase augmentation (NPA), 
which enables us to use good quality negative examples
by using the linguistic relationship (e.g., mutually exclusiveness) and 
the confusion between the categories (Fig.~\ref{fig:neg_aug}c).
It significantly improves the discriminative ability of the 
generated classifiers.

\subsubsection{Basic Training}
\label{sec:basic}
First, we describe the basic training strategy without 
NPA (Fig.~\ref{fig:neg_aug}b).
Training a Query-Adaptive R-CNN requires the phrases and their 
corresponding bounding boxes to be annotated.
For the $i$th image (we use one image as a minibatch), 
let us assume that $C_i$ phrases are associated with the image.
The $C_i$ phrases can be considered as the classes to train in the minibatch.
The labels $\mathbf{L}_i \in \{0,1\}^{C_i\times n_r}$ are assigned to 
the region proposals generated by RPN
(each of the dotted rectangles in Fig~\ref{fig:neg_aug}b);
a positive label is assigned if the box overlaps
the ground truth box by more than 0.5 in IoU and
negative labels are assigned to other RoIs under the assumption that 
all positive objects of $C_i$ classes are annotated
(i.e., regions without annotations are negative within the image).\footnote{Although this assumption 
is not always true for datasets such as Flickr30k Entities, 
it nonetheless works well for them because exceptions are rare.}
We then compute the classification loss by using the 
training labels and classification scores.\footnote{
Whereas Faster R-CNN uses the softmax cross entropy over the $C + 1$ (background) classes, 
where $C$ is the number of closed sets of a category,
we use the sigmoid cross entropy because the $C_i$ classes are not always 
mutually exclusive and a background class cannot be defined 
in the context of open-vocabulary object detection.
}
The loss in terms of RPN and bounding box regression is computed in the same way as Faster R-CNN~\cite{Ren2015}. 

\subsubsection{Negative Phrase Augmentation}
\label{sec:neg}
Here, we address the difficulty of using negative examples in the training of open-vocabulary object detection.
As shown in Fig.~\ref{fig:top}b, our generated classifier is not discriminative enough.
The reason is the scarcity of negative examples when using the training strategy described in Sec.~\ref{sec:basic}; 
e.g., the \texttt{horse} classifier is not learned with the \texttt{zebra} as a negative example
except for the rare case that both a \texttt{zebra} and a \texttt{horse} are in the same image.
Using hard negative examples has proven to be effective in the object detection 
to train a discriminative detector~\cite{Felzenszwalb2010,Girshick2014,Shrivastava2016}.
However, adding negative examples is usually not easy in the open-vocabulary setting,
because it is not guaranteed that a region without a positive label is negative.
For example, an object with the label \texttt{man} is not a negative of \texttt{person}
even though \texttt{person} is not annotated.
There are an infinite number of categories in open-vocabulary settings, 
which makes it difficult to exhaustively annotate all categories throughout the dataset.

How can we exploit hard examples that are guaranteed to be negative?
We can make use of the mutually exclusive relationship between categories:
e.g., an object with a \texttt{dog} label is negative for \texttt{cat} 
because \texttt{dog} and \texttt{cat} are mutually exclusive.
There are two ways we can add to a minibatch: 
add negative images (regions) or negative phrases.
Adding negative phrases (as in Fig.~\ref{fig:neg_aug}c) is generally better 
because it involves a much smaller additional training cost 
than adding images in terms of the both computational cost 
and GPU memory usage. 
In addition, to improve the discriminative ability of the classifier, 
we select only hard negative phrases by mining the confusing categories.
This approach, called {\it negative phrase augmentation (NPA)}, is a generic way of 
exploiting hard negative examples in open-vocabulary object detection and
leads to large improvements in accuracy, 
as we show in Sec.~\ref{sec:exp_ret}.

{\bf Confusion table.}
We create a confusion table that associates a category with its hard negative categories,
from which negative phrases are picked as illustrated in Fig.~\ref{fig:neg_aug}c.
To create the entry for category $c$,
we first generate the candidate list of hard negative categories 
by retrieving the top 500 scored objects 
from all objects in the validation set of Visual Genome~\cite{Krishna2016} (using $c$ as a query).
After that, we remove the mutually non-exclusive category relative to $c$ from the list.
Finally, we aggregate the list by category and assign a weight to each category.
Each of the registered entries becomes like {\small \texttt{dog:\{cat:0.5, horse:0.3, cow:0.2\}}}.
The weight corresponds to the probability of selecting the category in NPA,
which is computed based on the number of appearances and their ranks 
in the candidate list.\footnote{We compute the weight of each category 
as the sum of 500 minus the rank for all ranked results in the candidate lists normalized over all categories in order to sum to one.}

{\bf Removal of mutually non-exclusive phrases.}
To remove non-mutually exclusive phrases from the confusion table,
we use two approaches that estimate whether the two categories are mutually 
exclusive or not.
1) The first approach uses the {\it WordNet hierarchy}: 
if two categories have parent-child relationships in WordNet~\cite{Miller1995}, 
they are not mutually exclusive.
However, the converse is not necessarily true; e.g., \texttt{man} and \texttt{skier} are 
not mutually exclusive but do not have the parent-child relationship 
in the WordNet hierarchy.
2) As an alternative approach, we propose to use {\it Visual Genome annotation}:
if two categories co-occur more often in the Visual Genome dataset~\cite{Krishna2016}, 
these categories are considered to be not mutually exclusive.\footnote{ 
We set the ratio at 1\% of objects in either category.
For example, 
if there are 1000 objects with the \texttt{skier} label and 
20 of those objects are also annotated with \texttt{man} (20/1000=2\%), 
we consider that \texttt{skier} and \texttt{man} are not mutually exclusive.}
These two approaches are complementary, and they improve detection performance 
by removing the mutually non-exclusive words (see Sec.~\ref{sec:exp_ret}).

{\bf The training pipeline} with NPA is as follows:
\begin{enumerate}[(1)]
\setlength{\leftskip}{-7pt}
\setlength{\itemsep}{-4pt}
\setlength{\parsep}{0pt}
\item{\bf Update the confusion table:}
The confusion table is updated periodically (after every 10k iterations in our study).
Entries were created for categories 
that frequently appeared in 10k successive batches 
(or the whole training set if the size of the dataset is not large).
\item{\bf Add hard negative phrases:}
Negative phrases are added to each of the $C_i$ phrases in a minibatch.
We replace the name of the category in each phrase with its hard negative category 
(e.g., generate \texttt{a running woman} for \texttt{a running man}), 
where the category name is obtained by extracting nouns.
A negative phrase is randomly selected from the confusion table on the basis of the assigned probability.
\item{\bf Add losses:}
As illustrated in Fig.~\ref{fig:neg_aug}c,
we only add negative labels to the regions where a positive label is assigned to the original phrase.
The classification loss is computed only for the regions, which is added to the original loss.
\end{enumerate}


\section{Large-Scale Object Retrieval} 
\label{sec:ret}
Query-Adaptive R-CNN can be used for large-scale object retrieval and localization,
because it can be decomposed into a query-independent part and 
a query-dependent part, 
i.e., a region feature extractor and detector generator.
We follow the approach used in large-scale R-CNN~\cite{Hinami2016}, 
but we overcome its two critical drawbacks.
First, a large-scale R-CNN can only predict boxes included 
in the region proposals;
these are detected offline even though the query is unknown at the time; therefore,
to get high recall, a large number of object proposals should be used,
which is memory inefficient.
Instead, we generate a regressor as well as a classifier, 
which enables more accurate localization with fewer proposals.
Second, a large-scale R-CNN assumes that the classifier is given as a query, 
and learning a classifier requires many samples with bounding annotations.
We generate the classifier from a text query directly 
by using the detector generator of Query-Adaptive R-CNN.
The resulting system is able to retrieve and localize objects 
from a database with {\it one million images} in {\it less than one second}.

\textbf{Database indexing.}
For each image in the database,
the region feature extractor extracts 
region proposals and corresponding features.
We create an index for the region features in order to speed up the search.
For this, we use the IVFADC system~\cite{Jegou2011} in the manner described in \cite{Hinami2016}.

\textbf{Searching.}
Given a text query, the detector generator generates a linear classifier and bounding box regressor.
The regions with high classification scores are then retrieved from the database by making an IVFADC-based search.
Finally, the regressor is applied to the retrieved regions to obtain the accurately localized bounding boxes.

\begin{table*} 
\begin{center}
\scalebox{0.7}{
\begin{tabular}{l|c c c c c c c c|c}
\toprule
\multicolumn{1}{@{}l|}{Approach} & People & Clothing & Body & Animals & Vehicles & Instruments & Scene & Other & All \\
\midrule
\multicolumn{1}{@{}l|}{\small{\bf Non-scalable methods}} & & & & & & & & & \\
GroundeR~\cite{Rohrbach2016}                 & 61.00 & 38.12 & 10.33 & 62.55 & 68.75 & 36.42 & 58.18 & 29.08 & 47.81 \\
Multimodal compact bilinear~\cite{Fukui2016}  & - & - & - & - & - & - & - & - & 48.69 \\
PGN+QRN~\cite{Chen2017}                      & 75.08 & 55.90 & 20.27 & 73.36 & 68.95 & 45.68 & 65.27 & 38.80 & 60.21 \\
\midrule
\multicolumn{1}{@{}l|}{\small{\bf Non-scalable and joint localization methods}} & & & & & & & & & \\
Structured matching~\cite{Wang2016a}     & 57.89 & 34.61 & 15.87 & 55.98 & 52.25 & 23.46 & 34.22 & 26.23 & 42.08 \\
SPC+PPC~\cite{Cues}                      & 71.69 & 50.95 & 25.24 & 76.25 & 66.50 & 35.80 & 51.51 & 35.98 & 55.85 \\
QRC net~\cite{Chen2017}                  & 76.32 & 59.58 & 25.24 & {\bf 80.50} & {\bf 78.25} & 50.62 & 67.12 & 43.60 & 65.14 \\
\midrule
\multicolumn{1}{@{}l|}{\small{\bf Scalable methods}} & & & & & & & & & \\
Structure-preserving embedding~\cite{Wang} & - & - & - & - & - & - & - & - & 43.89 \\
CCA+Detector+Size+Color~\cite{Plummer2017} & 64.73 & 46.88 & 17.21 & 65.83 & 68.75 & 37.65 & 51.39 & 31.77 & 50.89 \\
\rowcolor{my-gray} {\bf Query-Adaptive R-CNN (proposed)}          & {\bf 78.17} & {\bf 61.99} & {\bf 35.25} & 74.41 & 76.16 & {\bf 56.69} & {\bf 68.07} & {\bf 47.42} & {\bf 65.21} \\
\bottomrule
\end{tabular}
}
\end{center}
\vspace{-2mm}
\caption{{\bf Phrase localization} accuracy on Flickr30k Entities dataset.}
\vspace{0mm}
\label{tab:phrase}
\end{table*}

\begin{table} 
\begin{center}
\scalebox{0.75}{
{\tabcolsep=1.8mm
\begin{tabular}{@{}l|l|c c c c c@{}}
\toprule
& & \multicolumn{5}{c}{{IoU}} \\ 
Architecture & Params & 0.5 & 0.6 & 0.7 & 0.8 & 0.9 \\
\midrule
 w/o regression & - & {\bf 65.21} & 53.19 & 35.70 & 14.32 & 1.88  \\
\midrule
 300--16(--4096) & {\bf 0.3M} & 64.14 & 57.66 & 48.22 & 33.04 & 9.29 \\
 300--64(--4096) & 1.1M & 63.87 & 57.43 & {\bf 49.05} & 33.84 & {\bf 10.55} \\
 300--256(--4096) & 4.3M & 63.84 & 57.70 & 48.71 & 33.87 & 10.05 \\
 300--1024(--4096) & 17M & 64.29 & {\bf 58.05} & 48.49 & {\bf 33.94} & 10.09 \\
 300(--256--4096) & 4.5M & 62.82 & 56.28 & 48.02 & 32.71 & 9.89 \\
 300--4096 & 1.2M & 63.23 & 56.92 & 48.17 & 32.66 & 9.20  \\
\bottomrule
\end{tabular}
}
}
\end{center}
\vspace{-2mm}
\caption{Comparison of various {\bf bounding box regressors} 
on Flickr30k Entities for different IoU thresholds. The number of parameters in $G_r$ is also shown.}
\vspace{0mm}
\label{tab:res_regressor}
\end{table} 

\section{Experiments} 
\subsection{Experimental Setup}
\textbf{Model:}
Query-Adaptive R-CNN is based on VGG16~\cite{Simonyan2015},
as in other work on phrase localization.
We first initialized the weights of the VGG and RPN by using
Faster R-CNN trained on Microsoft COCO~\cite{Lin2014};
the weights were then fine-tuned for each dataset of the evaluation.
In the training using Flickr30k Entities, we first pretrained the model
on the Visual Genome dataset using the object name annotations.
We used Adam~\cite{Kingma2015} with a learning rate starting
from 1e-5 and ran it for 200k iterations.

\textbf{Tasks and datasets:}
We evaluated our approaches on two tasks: 
phrase localization and open-vocabulary object detection and retrieval.
The \textbf{phrase localization task} was performed on the Flickr30k Entities dataset~\cite{Plummer2015}.
Given an image and a sentence that describes the image, 
the task was to localize region that corresponds to the phrase in a sentence.
Flickr30k datasets contain 44,518 unique phrases, where 
the number of words of each phrase is 1--8 (2.1 words on average).
We followed the evaluation protocol of \cite{Plummer2015}.
We did not use Flickr30k Entities for the retrieval task 
because the dataset is not exhaustively annotated
(e.g., not all men appearing in the dataset are annotated with 
\texttt{man}), 
which makes it difficult to evaluate with a retrieval metric such as
AP, as discussed in Plummer et al.~\cite{Plummer2017}.
Although we cannot evaluate the retrieval performance directly 
on the phrase localization task,
we can make comparisons with other approaches and 
show that our method can handle a wide variety of phrases.

The \textbf{open-vocabulary object detection and retrieval task}
was evaluated in the same way as the standard object detection task.
The difference was the assumption that we do not know the target category 
at training time in open-vocabulary settings;
i.e., the method does not tune in to a specific category, 
unlike the standard object detection task.
We used the Visual Genome dataset~\cite{Krishna2016} 
and selected the 100 most frequently object categories as queries among its 100k or so categories.\footnote{Since 
the WordNet synset ID is assigned to each object, 
we add objects with labels of hyponyms as positives
(e.g., \texttt{man} is positive for the \texttt{person} category).}
\footnote{We exclude the background (e.g., \texttt{grass}, \texttt{sky}, 
\texttt{field}), multiple objects (e.g., \texttt{people}, \texttt{leaves}), 
and ambiguous categories (e.g, \texttt{top}, \texttt{line}). }
We split the dataset into training, validation, and test sets following \cite{Johnson2015}.
We also evaluated our approaches on the PASCAL VOC 2007 dataset, 
which is a widely used dataset for object detection.\footnote{
We used the model trained on Visual Genome even for the evaluation on the PASCAL dataset
because of the assumption that the target category is unknown.
}
As metrics, we used top-k precision and average precision (AP),
computed from the region-level ranked list as in the standard object detection task.\footnote{
We did not separately evaluate the detection and retrieval tasks 
because both can be evaluated with the same metric.}

\begin{figure*}[t] 
\begin{center}
\includegraphics[width=1.00\linewidth]{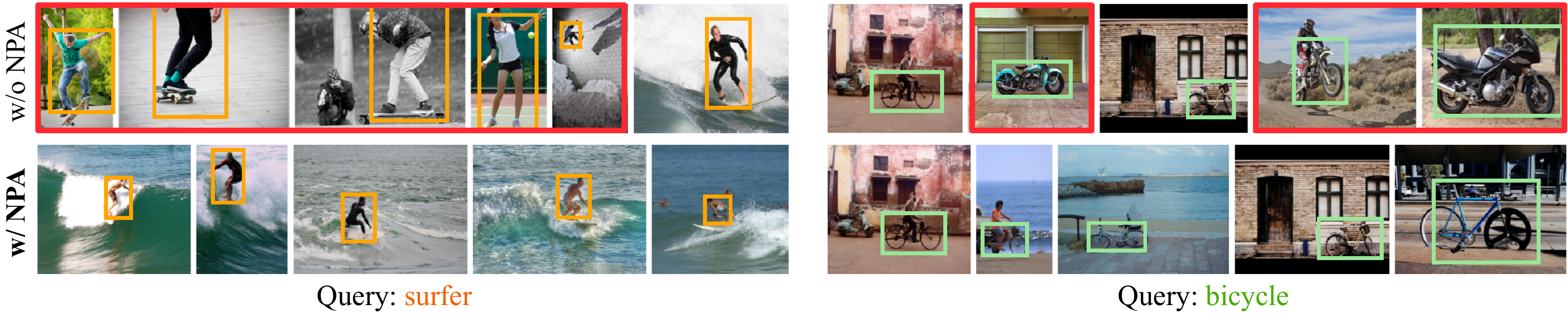}
\vspace{-6mm}
\caption{Qualitative results with and without NPA.
Top-k retrieved results for two queries are shown (sorted by rank) and false alarms are depicted with a red border.}
\vspace{0mm}
\label{fig:aug_qual} 
\end{center} 
\end{figure*}

\subsection{Phrase localization}
\label{sec:exp_ph}
\textbf{Comparison with state-of-the-art.}
We compared our method with state-of-the-art methods on
the Flickr30k Entities phrase localization task.
We categorized the methods into two types, i.e., non-scalable
and scalable methods (Tab.~\ref{tab:phrase}).
1) {\it Non-scalable methods} 
cannot be used for large-scale retrieval because 
their query-dependent components are too complex to 
process a large amount of images online, and
2) {\it Scalable methods} can be used for large-scale retrieval
because their query-dependent components are easy to scale up
(e.g., the $L_2$ distance computation);
these include common subspace-based approaches such as CCA.
Our method also belongs to the scalable category.
We used a simple model without a regressor and NPA 
in the experiments.

Table~\ref{tab:phrase} compares Query-Adaptive R-CNN  
with the state-of-the-art methods.
Our model achieved {\it 65.21\%} in accuracy and 
outperformed all of the previous state-of-the-art models including
the non-scalable or joint localization methods.
Moreover, it significantly outperformed the scalable methods, 
which suggests the approach of predicting the classifier is better 
than learning a common subspace for the open-vocabulary detection problem.

\begin{figure*}[t] 
\begin{center}
\includegraphics[width=1.00\linewidth]{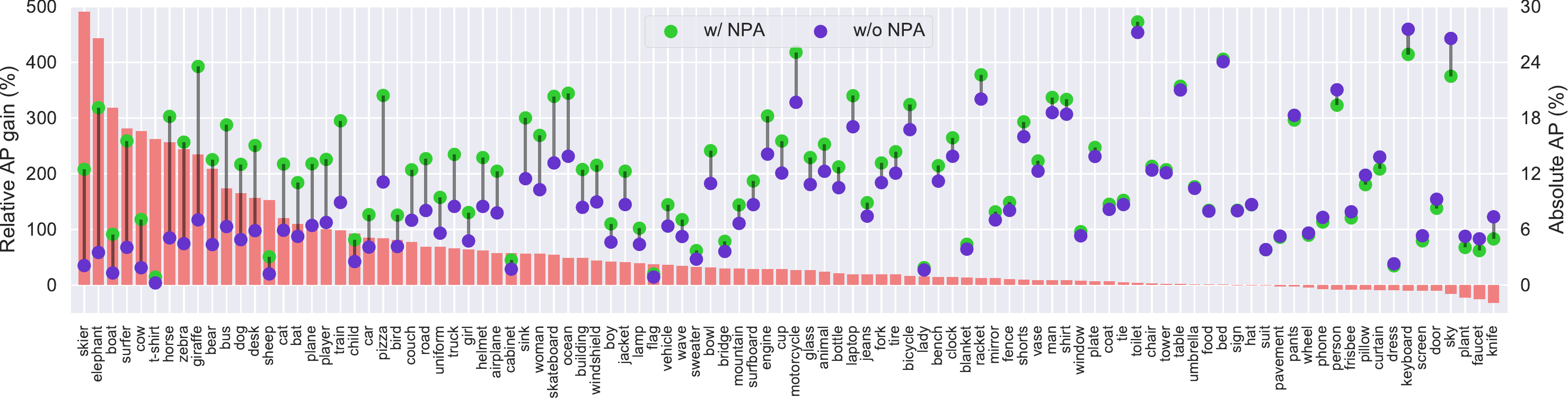}
\vspace{-7mm}
\caption{AP gain by \textbf{negative phrase augmentation} (NPA)
for individual queries. The bars show the relative AP gain and 
points shows the absolute AP with and without NPA.}
\vspace{0mm}
\label{fig:res_aug} 
\end{center} 
\end{figure*}

{\bf Bounding box regressor.}
To demonstrate the effectiveness of the bounding box regressor for precise localization, 
we conducted evaluations with the regressor at different IoU thresholds.
As explained in Sec.~\ref{sec:qrcn_arch},
the regressor was generated using $G_r$,
which transformed 300-d text embeddings $x$ into 4096-d regressor weights 
$\mathbf{w}^r_x$, $\mathbf{w}^r_y$, $\mathbf{w}^r_w$, and $\mathbf{w}^r_h$.
We compared three network architectures for $G_r$:
1) \texttt{300-n(-4096)} MLP having a hidden layer with $n$ units that
is shared across the four outputs,
2) \texttt{300(-n-4096)} MLP having a hidden layer that is not shared, 
and 3) \texttt{300(-4096)} linear transformation (without a hidden layer).

Table~\ref{tab:res_regressor} shows the results with and 
without regressor.
The regressor significantly improved the accuracy 
with high IoU thresholds,
which demonstrates that the regressor improved the localization accuracy.
In addition, the accuracy did not decrease as a result of sharing the hidden layer 
or reducing the number of units in the hidden layer.
This suggests that the regressor lies in a very low-dimensional manifold 
because the regressor for one concept can be shared by many concepts 
(e.g., the \texttt{person} regressor can be used for \texttt{man}, \texttt{woman}, \texttt{girl}, \texttt{boy}, etc.).
The number of parameters was significantly reduced by these tricks,
to even fewer than in the linear transformation. 
The accuracy slightly decreased with a threshold of 0.5, 
because the regressor was not learned properly for the 
categories that did not frequently appear in the training data.

\begin{table} 
\begin{center}
\vspace{-2mm}
\scalebox{0.80}{
{\tabcolsep=1.8mm
\begin{tabular}{@{}p{1.5cm}|c c c|c c c|c@{}}
\toprule
& & & & \multicolumn{3}{c|}{{\bf Visual Genome}} & \multicolumn{1}{c}{{\bf VOC}} \\ 
&\small{NPA}&\small{WN}&\small{VG}& \small{mAP} & \scriptsize{PR@10} & \scriptsize{PR@100} & \small{mAP} \\
\midrule
CCA  & & &           & 3.18  & 20.40 & 15.64 & 28.23 \\
\midrule
\multirow{5}{*}{\shortstack{{\bf Query-}\\{\bf Adaptive} \\{\bf R-CNN}} }& &  & & 9.15  & 52.60 & 36.85 & 29.14 \\
&\cmark &        &        & 10.90 & 60.10 & 43.21 & 36.74 \\
&\cmark & \cmark &        & 11.53 & 61.80 & 45.91 & 37.07 \\
&\cmark &        & \cmark & 11.65 & 65.40 & 46.85 & 41.32 \\
&\cmark & \cmark & \cmark & {\bf 12.19} & {\bf 65.70} & {\bf 48.45} & {\bf 42.81} \\
\bottomrule
\end{tabular}
}
}
\end{center}
\vspace{-2mm}
\caption{\textbf{Open-vocabulary object detection} performance 
on Visual Genome and PASCAL VOC 2007 datasets. WN and VG are
the strategies to remove mutually non-exclusive phrases.}
\vspace{-2mm}
\label{tab:open}
\end{table}

\begin{table}[t] 
\begin{center}
\scalebox{0.78}{
{\tabcolsep=1.8mm
\begin{tabular}{@{}l|l c c c|l c c c@{}}
\toprule
Query & \multicolumn{4}{l|}{Most confusing class}  & \multicolumn{4}{l}{2nd most confusing class} \\
\midrule
girl & man & 19 & $\to$ & 3 & boy & 4 & $\to$ & 2\\
skateboard & surfboard & 12 & $\to$ & 0 & snowboard & 11 & $\to$ & 0\\
train & bus & 17 & $\to$ & 1 & oven & 3 & $\to$ & 0\\
helmet & hat & 18 & $\to$ & 1 & cap & 6 & $\to$ & 4\\
elephant & bear & 14 & $\to$ & 0 & horse & 6 & $\to$ & 0\\
\bottomrule
\end{tabular}
}
}
\end{center}
\vspace{-2mm}
\caption{Number of false alarms in top 100 results for five queries ({\bf w/o NPA $\to$ w/ NPA}). 
The top 2 confusing categories are shown for each query.}
\vspace{-3mm}
\label{tab:confuse}
\end{table}

\begin{figure*}[t] 
\vspace{-1mm}
\begin{center}
\includegraphics[width=1.00\linewidth]{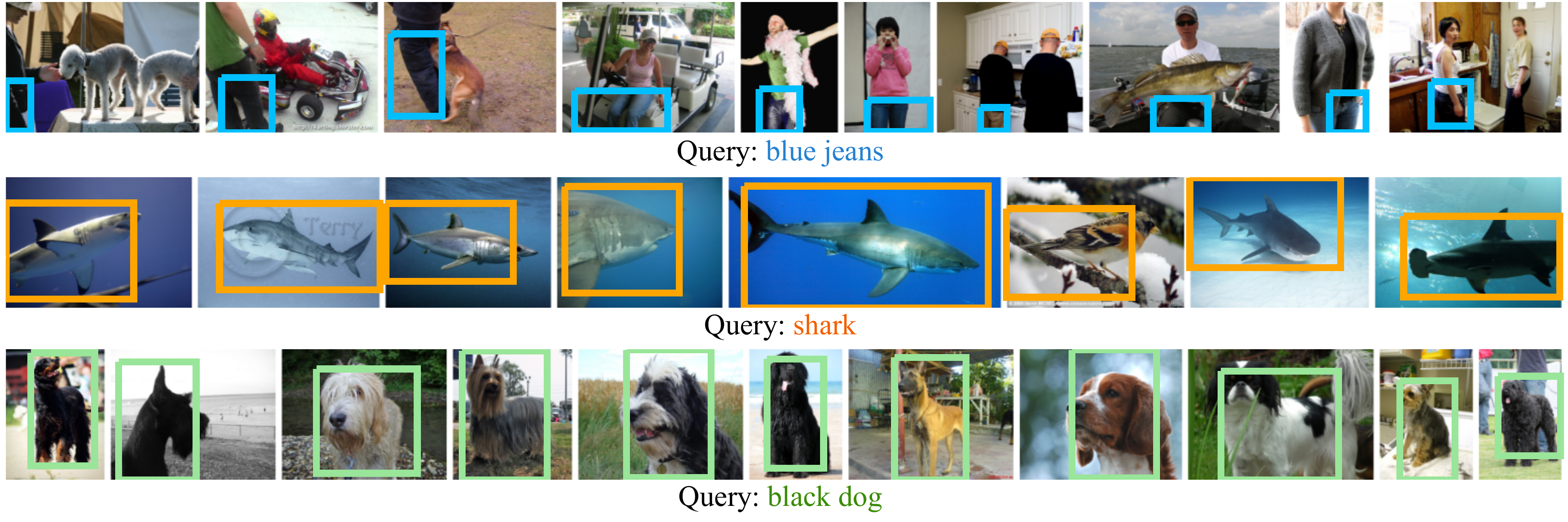}
\vspace{-8mm}
\caption{Retrievals from one million images.
Top-k results for three queries are shown. 
}
\vspace{0mm}
\label{fig:large_qual} 
\end{center} 
\end{figure*}

\subsection{Open-Vocabulary Object Retrieval}
\label{sec:exp_ret}
\textbf{Main comparison.} 
Open-vocabulary object detection and retrieval
is a much more difficult task than phrase localization,
because we do not know how many objects are present in an image.
We used NPA to train our model.
As explained in Sec.~\ref{sec:neg},
we used two strategies, {\it Visual Genome annotation (VG)} and 
{\it WordNet hierarchy (WN)}, to remove mutually non-exclusive phrases 
from the confusion table.
As a baseline, we compared with region-based CCA~\cite{Plummer2017},
which is scalable and shown to be effective for phrase localization;
for a fair comparison, the subspace was learned using the same dataset as ours.
An approximate search was not used to evaluate the actual performance
at open-vocabulary object detection.

Table~\ref{tab:open} compares different training strategies.
NPA significantly improved the performance: 
{\it more than 25\% relative improvement} for all metrics.
Removing mutually non-exclusive words also 
contributed the performance: WN and VG both improved performance
(5.8\% and 6.9\% relative AP gain, respectively).
Performance improved even further by combining them 
(11.8\% relative AP gain), which shows they are complementary.
AP was much improved by NPA for the PASCAL dataset as well 
(47\% relative gain).
However, the performance was still much poorer than 
those of the state-of-the-art object detection 
methods~\cite{JosephRedmon2016,Ren2015}, 
which suggests that there is a large gap 
between open-vocabulary and closed-vocabulary object detection.

\textbf{Detailed results of NPA.}
To investigate the effect of NPA, we show the AP with and without NPA 
for individual categories in Figure~\ref{fig:res_aug}, 
which are sorted by relative AP improvement.
It shows that AP improved especially for 
animals (\texttt{elephant}, \texttt{cow}, \texttt{horse}, etc.) and
person (\texttt{skier}, \texttt{surfer}, \texttt{girl}),
which are visually similar within the same upper category.
Table~\ref{tab:confuse} shows the most confused category and 
its total count in the top 100 search results for each query,
which shows what concept is confusing for each query
and how much the confusion is reduced by NPA.\footnote{For 
each query, we scored all the objects in the Visual Genome testing set 
and counted the false alarms in the top 100 scored objects.}
This shows that visually similar categories resulted in false positive without NPA, 
while their number was suppressed by training with NPA.
The reason is that these confusing categories were added for negative 
phrases in NPA, and the network learned to reject them.
Figure~\ref{fig:aug_qual} shows the qualitative search results  
for each query with and without NPA (and CCA as a baseline),
which also showed that NPA can discriminate confusing categories
(e.g., \texttt{horse} and \texttt{zebra}).
These results clearly demonstrate that NPA significantly 
improves the discriminative ability of classifiers by adding 
hard negative categories.
\newline

\begin{table}  
\begin{center} 
\vspace{-1mm}
\scalebox{0.73}{
{\tabcolsep=1.6mm
\begin{tabular}{@{}l|c c c c c c@{}} \toprule
Database size & 10K  & 50K  & 100K  & 500K & 1M \\
\midrule                                   
Time (ms) & 183$\pm$16 & 196$\pm$21 & 242$\pm$28  & 314$\pm$90  & 484$\pm$165  \\
Memory (GB) & 0.46 & 1.23 & 2.19 & 9.87 & 19.47\\
\bottomrule
\end{tabular} 
}
}
\end{center} 
\vspace{-3mm}
\caption{Speed/memory in {\bf large-scale} experiments.}
\vspace{0mm}
\label{tab:res_large} 
\end{table} 

\textbf{Large-scale experiments.}
Finally, we evaluated the scalability of our method on a large image database.
We used one million images from the ILSVRC 2012 training set for 
this evaluation. 
Table~\ref{tab:res_large} show the speed and memory.  
The mean and standard deviation of speed are computed over 20 queries in PASCAL VOC dataset.
Our system 
could retrieve objects from one million images in around 0.5 seconds.
We did not evaluate accuracy because 
there is no such large dataset with bounding box annotations.\footnote{
adding distractors would also be difficult, because we 
cannot guarantee that relevant objects are not in the images.}
Figure~\ref{fig:large_qual} shows the retrieval results from one million images, 
which demonstrates that our system can accurately retrieve and localize objects 
from a very large-scale database.

\section{Conclusion} 
Query-Adaptive R-CNN is a simple yet strong framework for 
open-vocabulary object detection and retrieval. 
It achieves state-of-the-art performance on the Flickr30k 
phrase localization benchmark and
it can be used for large-scale object retrieval by textual query. 
In addition, its retrieval accuracy can be further increased 
by using a novel training strategy called negative phrase augmentation (NPA) 
that appropriately selects hard negative examples by using 
their linguistic relationship and confusion between categories.
This simple and generic approach significantly improves the 
discriminative ability of the generated classifier.

{\bf Acknowledgements:}
This work was supported by JST CREST JPMJCR1686 
and JSPS KAKENHI 17J08378.
\clearpage
\bibliography{QRCN}
\bibliographystyle{acl_natbib_nourl}

\onecolumn
\section*{\Large{Supplementary Material}}
\setcounter{section}{0}
\section{Detailed Analysis on Phrase Localization} 
\subsection{With or Without Regression}
\begin{figure*}[h] 
\begin{center}
\includegraphics[width=1.00\linewidth]{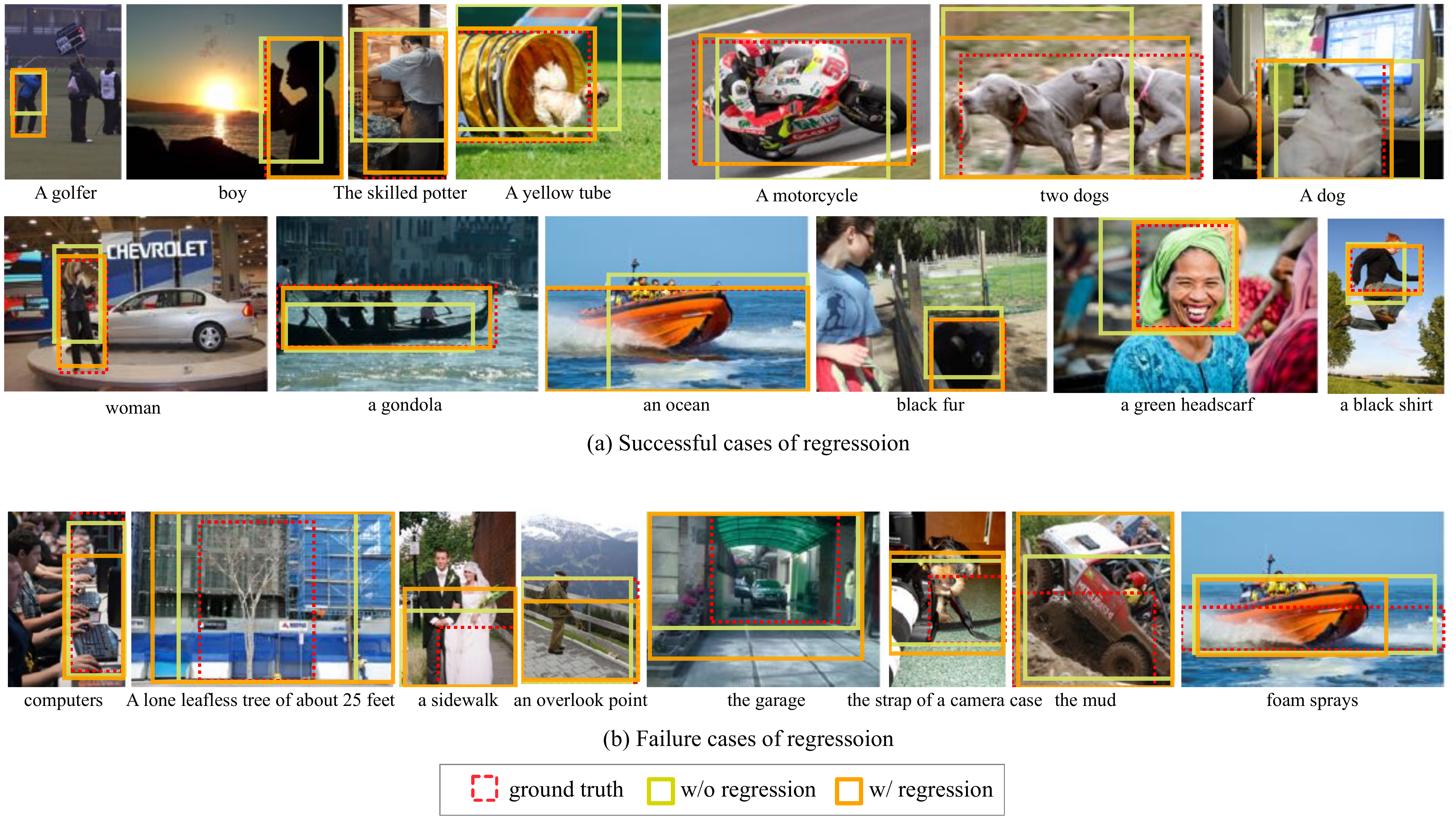}
\caption{Phrase localization results with and without the bounding box regression.
We visualize the ground truth bounding box, the result without the regression, 
and the result with NPA in red, yellow, and orange, respectively.}
\label{fig:reg_ph_qual} 
\end{center} 
\end{figure*}

Figure~\ref{fig:reg_ph_qual} compares the results with and without bounding box regression.
We use the model of \texttt{300-16(-4096)} to generate the regressor
(explained in our paper in Sec.5.2).
Figure~\ref{fig:reg_ph_qual}a shows the successful cases.
The regression is effective especially for
the frequently appeared categories in training data such as \texttt{person} and \texttt{dog}
because the accurate regressor can be learned by using many examples.
The regression was succeeded for several uncommon categories such as \texttt{potter} and \texttt{gondola};
the reason is that regressor can be shared with other common categories,
e.g., \texttt{person} and \texttt{boat} regressor can be used for \texttt{potter} and \texttt{gondola}, respectively.
Figure~\ref{fig:reg_ph_qual}b shows the failure cases,
which include the categories with ambiguous boundary (e.g., \texttt{sidewalk} and \texttt{mud}).
The regressor does not work for such categories. 
In addition, if the category is not frequently appeared in training data,
the regressor moves the bounding box into the wrong direction.
Future work includes automatically determining whether 
to perform bounding box regression or not.

\clearpage
\subsection{With or Without Negative Phrase Augmentation} 
\begin{figure*}[h] 
\begin{center}
\includegraphics[width=1.00\linewidth]{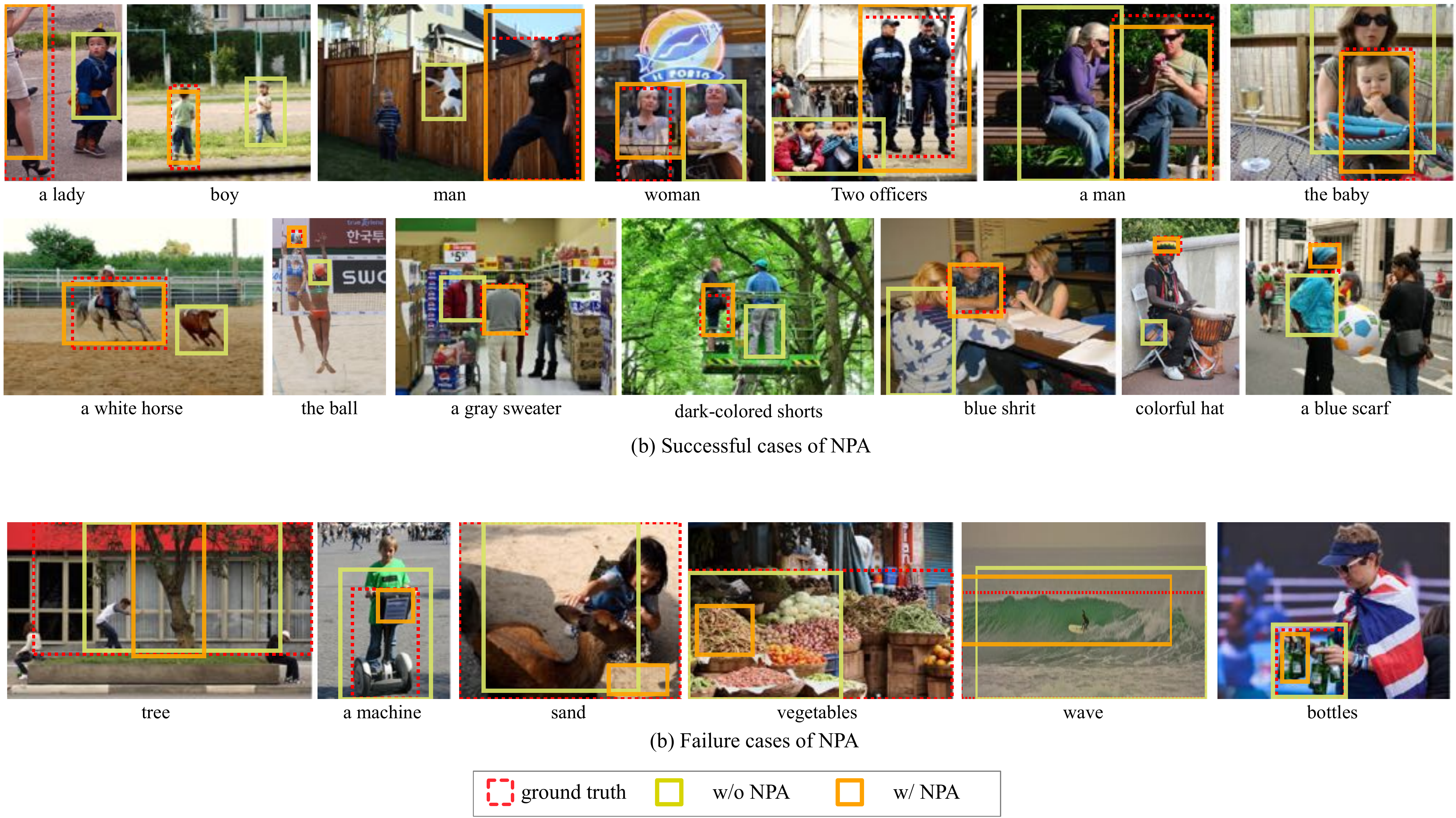}
\caption{Phrase localization results with and without bounding box regression. We visualize ground truth bounding box, 
the result without NPA, and the result with NPA in red, yellow, and orange, respectively.}
\label{fig:aug_ph_qual} 
\end{center} 
\end{figure*}

Table~\ref{tab:aug_ph_quan} shows the phrase localization performance with and without negative phrase augmentation (NPA).
It shows that the phrase localization performance is not improved by training with NPA.
As explained in our paper, it is due to the difference between the phrase localization and object detection tasks;
phrase localization assumes there is only one relevant object in the image while object detection places no assumption on the number of objects.
Because of this, in the phrase localization task, we can benefit from NPA only 
when confusing objects appear in the single image.
Figure~\ref{fig:aug_ph_qual}a shows such cases: e.g.,
when two persons appear in the same image, 
the method with NPA can select the appropriate person that is relevant to the query.
However, since such cases are rare in the Flickr30k Entities dataset, NPA does not 
contribute to performance.
Figure~\ref{fig:aug_ph_qual}b shows the failure cases of NPA.
The method with NPA tends to predict the small bounding box in which 
other objects do not appear. 
The reason is that NPA cannot handle highly overlapped objects appropriately.
For example, in the third example in Fig.~\ref{fig:aug_ph_qual}, 
the \texttt{sand} region may have high scores for the \texttt{deer}. 
If there are many such cases in the validation set, 
the \texttt{sand} is added to hard negative phrase for the \texttt{deer}.
The \texttt{sand} classifier thus predicts low score to the regions that are overlapped with the \texttt{deer}.
Therefore, the method with NPA tends to predict the small box that contains only the (part of) relevant object.
This is the limitation of NPA and causes the accuracy decrease in phrase localization task.

\begin{table*}[h]
\begin{center} 
\scalebox{0.90}{
\begin{tabular}{l|c c c c c c c c|c} 
\toprule
Method & People & Clothing & Body & Animals & Vehicles & Inst. & Scene & Other & All \\
\midrule
w/o NPA & {\bf 78.17} & {\bf 61.99} & {\bf 35.25} & 74.41 & {\bf 76.16} & 56.69 & 68.07 & {\bf 47.42} & {\bf 65.21} \\
w/ NPA & 77.13 & 60.06 & 33.86 & {\bf 76.76} & 73.55 & {\bf 58.60} & {\bf 68.94} & 45.28 & 64.09\\
\bottomrule
\end{tabular} 
}
\end{center} 
\caption{Comparison of Flickr30k Entities phrase localization performance with and without NPA.}
\label{tab:aug_ph_quan} 
\end{table*}

\clearpage
\subsection{Ablation Studies} 
We here present the detailed analysis of our approach 
on the Flickr30k Entities phrase localization task
and quantify our architectural design decisions.
For the simplicity, in the comparison of the region proposal and 
text embedding, we used the pretrained Faster R-CNN model trained on the 
COCO object detection and finetuned it for phrase localization task.
The bounding box regression and NPA are not used in this experiments.

{\bf Pretraining.}
Table~\ref{tab:pre} compares three pretrained models trained on 
1) ImageNet classification,
2) PASCAL, and 3) COCO object detection.
In addition, we pretrain the whole model including detector generator
using Visual Genome dataset after initial pretraining of 1)--3).
The results show that there is more than 4\% difference in accuracy
between simply using ImageNet pretrained model 
and pretraining on COCO and Visual Genome.
Since Flickr30k Entities dataset does not contain many training examples
for each object category, pretraining Faster R-CNN 
with large object detection datasets is important.
Training on the Visual Genome dataset further improves the performance because it contains 
a much larger number of categories than COCO dataset and detector generator is also pretrained on such rich data.

\begin{table*}[h] 
\begin{center}
\scalebox{0.85}{
\begin{tabular}{@{}l|c|c c c c c c c c|c@{}}
\toprule
Pretrained model& VG pretrain? & People & Clothing & Body & Animals & Vehicles & Inst. & Scene & Other & All \\
\midrule
ImageNet    &        & 74.98 & 57.34 & 28.12 & 71.88 & 70.93 & 50.32 & 67.45 & 40.34 & 60.97 \\
ImageNet    & \cmark & 76.30 & 58.30 & 27.72 & 74.61 & 69.19 & 56.06 & {\bf 69.07} & 43.93 & 62.76 \\
PASCAL      &        & 75.87 & 58.00 & 30.69 & 74.80 & 73.26 & {\bf 59.87} & 66.52 & 42.58 & 62.19 \\
PASCAL      & \cmark & 77.06 & 61.51 & 34.06 & 77.15 & 69.48 & 57.96 & 67.95 & 46.72 & 64.44 \\
COCO        &        & 77.26 & 60.19 & 33.86 & {\bf 75.78} & 75.29 & 56.69 & 66.83 & 46.01 & 64.08 \\
COCO        & \cmark & {\bf 78.17} & {\bf 61.99} & {\bf 35.25} & 74.41 & {\bf 76.16} & 56.69 & 68.07 & {\bf 47.42} & {\bf 65.21} \\
\bottomrule
\end{tabular}
}
\end{center}
\caption{Comparison of different {\bf pre-training} strategies on Flickr30k Entities phrase localization.}
\label{tab:pre}
\end{table*}

{\bf Region proposal.}
Table~\ref{tab:region} compares three region proposal approaches:
1) selective search~\cite{Uijlings2013},
2) region proposal network (RPN) trained on COCO dataset, 
which is frozen during the training of phrase localization,
and 3) RPN finetuned on phrase localization task.
The number of regions is 2000 for the selective search 
following~\cite{Girshick2014} and 300 for the RPN following~\cite{Ren2015}.
In addition, we compared two region sampling strategies: 
random sampling used in~\cite{Girshick2015,Ren2015}
and online hard example mining (OHEM)~\cite{Shrivastava2016}.
The results show that the RPN finetuned for phrase localization task 
generates much higher quality region proposals than others
(12.41\% increase in accuracy compared to the selective search), 
which demonstrates that learning region proposals play an important role 
in the phrase localization.
OHEM further improved the accuracy by 1.56\%.

\begin{table*}[h] 
\begin{center}
\scalebox{0.85}{
\begin{tabular}{@{}l|c|c c c c c c c c|c@{}}
\toprule
Region proposal & OHEM? & People & Clothing & Body & Animals & Vehicles & Inst. & Scene & Other & All \\
\midrule
Selective search          &        & 60.65 & 44.55 & 23.96 & 65.04 & 68.90 & 36.94 & 55.71 & 34.43 & 50.11 \\
RPN (COCO pretrained)     &        & 71.29 & 44.82 & 17.23 & 70.90 & 67.44 & 42.04 & 63.23 & 38.26 & 55.94 \\
RPN (COCO pretrained)     & \cmark & 72.12 & 42.62 & 16.24 & 71.88 & 67.15 & 44.59 & 65.09 & 36.45 & 55.71 \\
RPN (Flickr30k finetuned) &        & 75.90 & 58.74 & 28.32 & 72.66 & 73.55 & 55.41 & 65.34 & 44.88 & 62.52 \\
RPN (Flickr30k finetuned) & \cmark & {\bf 77.26} & {\bf 60.19} & {\bf 33.86} & {\bf 75.78} & {\bf 75.29} & {\bf 56.69} & {\bf 66.83} & {\bf 46.01} & {\bf 64.08} \\
\bottomrule
\end{tabular}
}
\end{center}
\caption{Comparison of different {\bf region proposals} and region sampling strategies on Flickr30k Entities phrase localization.}
\label{tab:region}
\end{table*}

{\bf Text embedding.}
Table~\ref{tab:text} compares five text embedding vectors: 
1) Word2Vec~\cite{Mikolov2013} trained on Google News dataset\footnote{https://code.google.com/archive/p/word2vec/}, 
which is used in our paper, 
2) Word2Vec trained on Flickr tags\footnote{the model is provided by the author of \cite{Dong2016}}~\cite{Li2015a},
3) Hybrid Gaussian-Laplacian mixture model (HGLMM)~\cite{Klein2015}, 
which is used in~\cite{Plummer2015,Plummer2017,Wang}, 
4) Skip-thought vector (combine-skip model)\footnote{We use the implementation and pre-trained model provided in https://github.com/ryankiros/skip-thoughts}~\cite{Kiros2015}, 
and 5) Long-short term memory (LSTM) that encodes a phrase into a vector in the manner described 
in~\cite{Chen2017,Rohrbach2016}, which is learned
jointly with other components of Query-Adaptive R-CNN.
The second column of Table~\ref{tab:text} shows the dimension of the text embedding vector.
This result shows that the performance is not much affected 
by the choice of the text embedding.
The mean pooling of Word2Vec performs the best despite its simplicity.

\begin{table*}[h] 
\begin{center}
\scalebox{0.85}{
\begin{tabular}{@{}l|c|c c c c c c c c|c@{}}
\toprule
Text embedding & dim & People & Clothing & Body & Animals & Vehicles & Inst. & Scene & Other & All \\
\midrule
Word2Vec avg.               & 300   & {\bf 77.26} & 60.19 & 33.86 & 75.78 & 75.29 & 56.69 & 66.83 & {\bf 46.01} & {\bf 64.08} \\
Word2Vec avg. (Flickr tags) & 300   & 75.36 & 60.19 & 31.88 & 75.00 & {\bf 78.78} & 55.41 & {\bf 68.39} & 44.64 & 63.19 \\
HGLMM                  & 15000 & {\bf 77.26} & {\bf 61.34} & 32.28 & 75.00 & 68.31 & {\bf 63.06} & 67.33 & 45.25 & 63.96 \\
Skip-thought           & 4800  & 77.06 & 59.89 & {\bf 34.65} & {\bf 79.88} & 73.55 & 57.32 & 68.01 & 45.28 & 64.06 \\
LSTM                   & 1000  & 75.45 & 58.96 & 28.71 & 74.61 & 75.58 & 56.05 & 66.71 & 29.23 & 62.36 \\
\bottomrule
\end{tabular}
}
\end{center}
\caption{Comparison of different {\bf text embedding} on Flickr30k Entities phrase localization. }
\label{tab:text}
\end{table*}

\clearpage
\section{Additional Examples of Negative Phrase Augmentation} 
\begin{figure*}[h] 
\begin{center}
\includegraphics[width=1.00\linewidth]{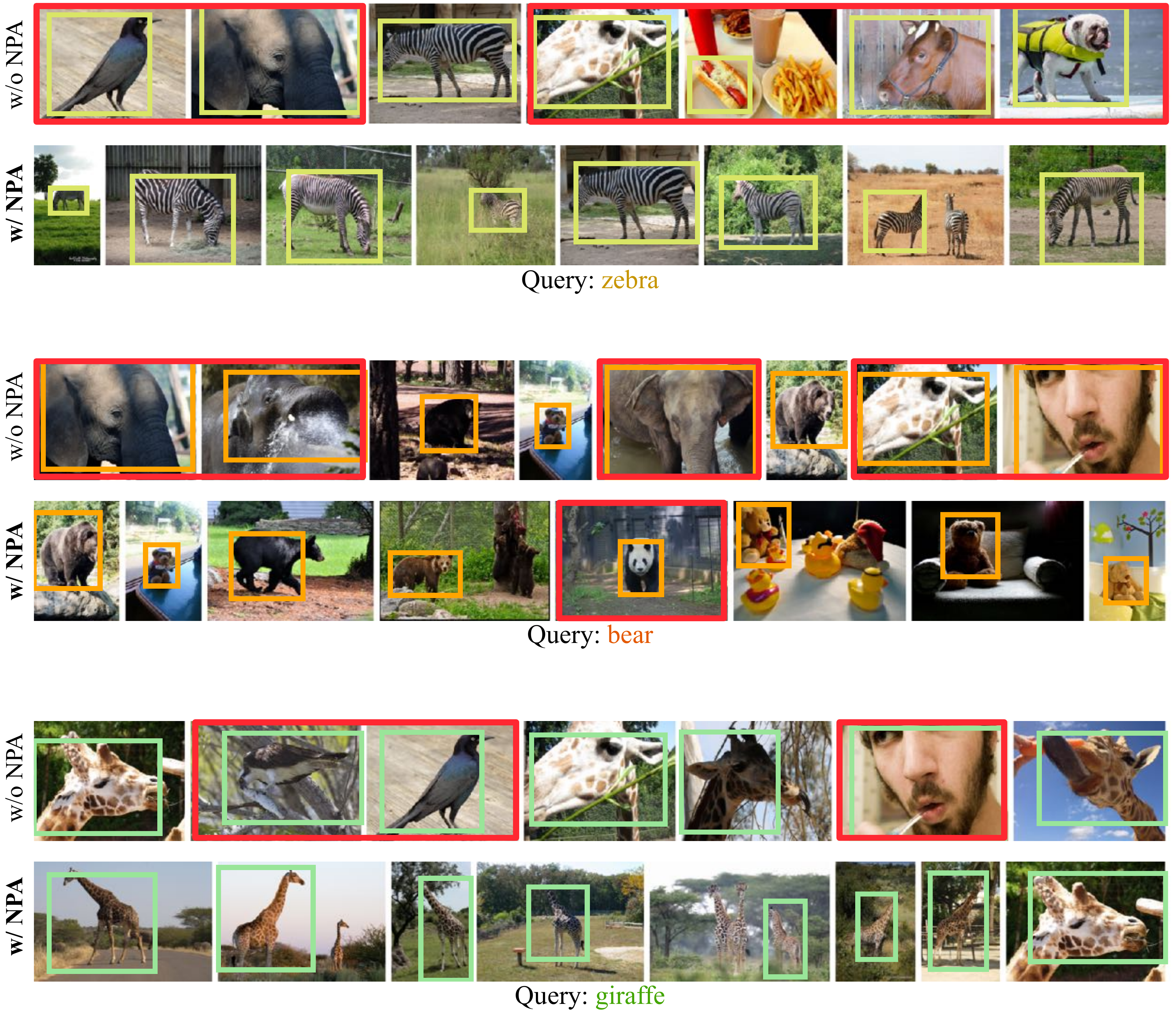}
\caption{Qualitative results with and without NPA.
Top-ranked retrieved results are shown and false alarms are depicted with red border.}
\label{fig:neg_qual1} 
\end{center} 
\end{figure*}

\begin{figure*}[t] 
\begin{center}
\includegraphics[width=1.00\linewidth]{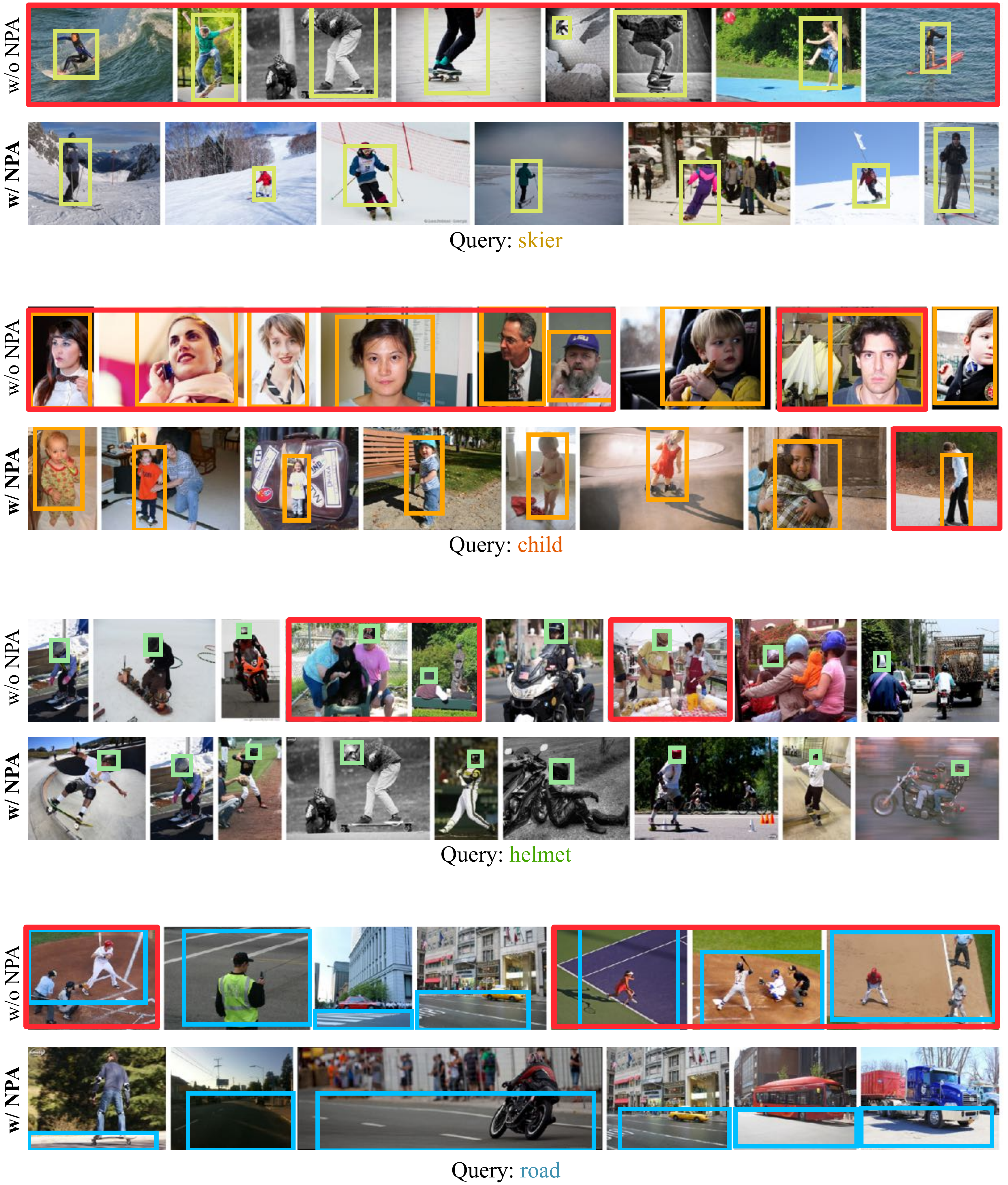}
\caption{Qualitative results with and without NPA.
Top-ranked retrieved results are shown and false alarms are depicted with red border.}
\label{fig:neg_qual2} 
\end{center} 
\end{figure*}

\begin{figure*}[t] 
\begin{center}
\includegraphics[width=1.00\linewidth]{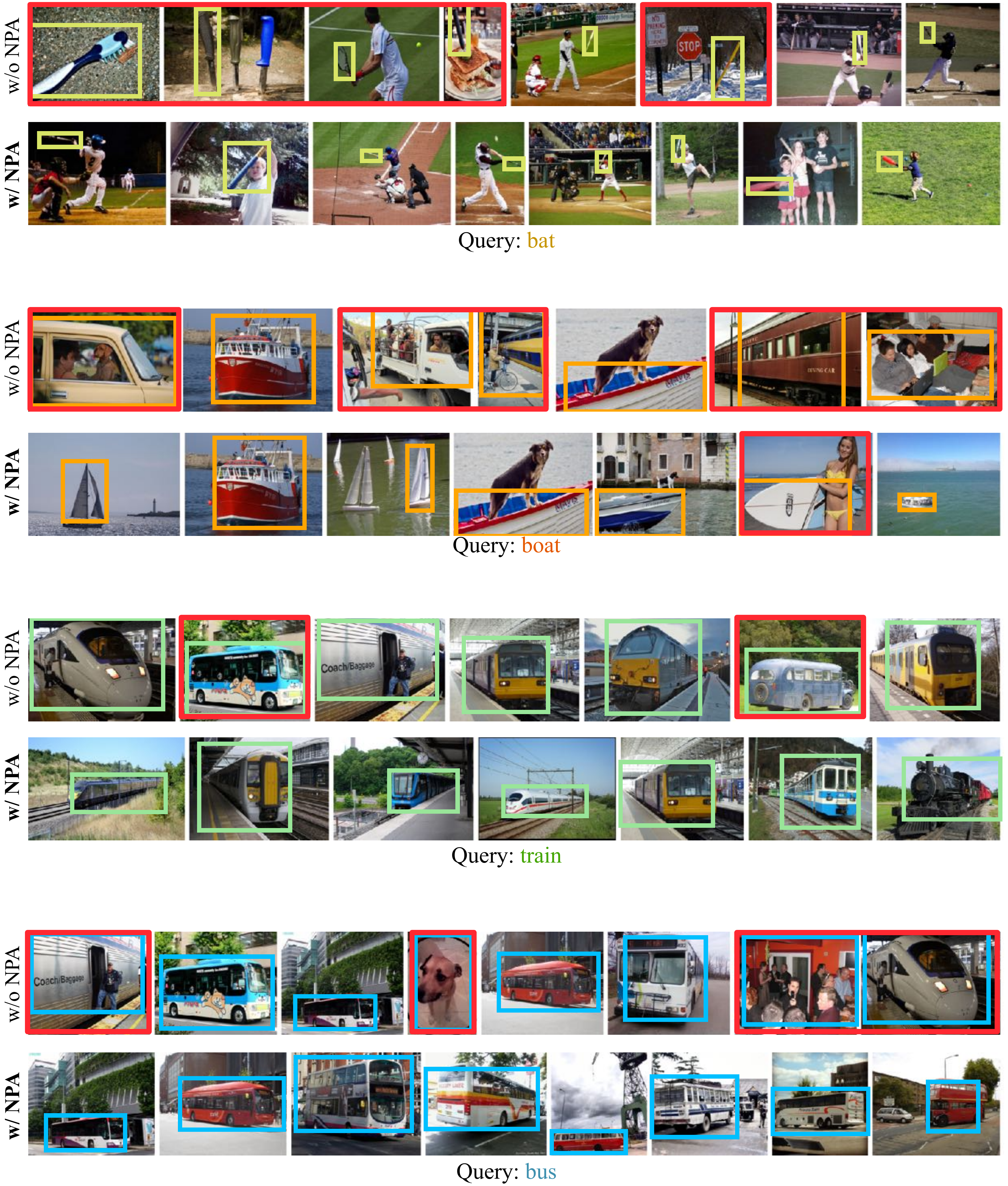}
\caption{Qualitative results with and without NPA.
Top-ranked retrieved results are shown and false alarms are depicted with red border.}
\label{fig:neg_qual3} 
\end{center} 
\end{figure*}

Figure~\ref{fig:neg_qual1}, \ref{fig:neg_qual2}, and \ref{fig:neg_qual3} 
show additional examples of the negative phrase augmentation 
(corresponds to Fig.~4 in our paper).
There are many false alarms between the confusing categories 
such as the animal 
(\texttt{zebra}, \texttt{bear}, and \texttt{giraffe}),
person (\texttt{skier} and \texttt{child}),
and vehicle (\texttt{boat}, \texttt{train}, and \texttt{bus}) without NPA,
which are successfully discarded by training with NPA.


\clearpage
\section{Additional Examples of Open-Vocabulary Object Retrieval and Localization} 
Figure~\ref{fig:large_qual} shows the additional examples of 
object retrieval and localization (corresponds to Fig.~6 in our paper).
Instead of the ILSVRC dataset used in our paper,
we here used the Microsoft COCO dataset~\cite{Lin2014} 
(40504 images from the validation set) 
that contains a wider variety of concepts.
These results demonstrate that our system can accurately 
search the wide variety of objects specified by the natural language query.

\begin{figure*}[h] 
\begin{center}
\includegraphics[width=1.00\linewidth]{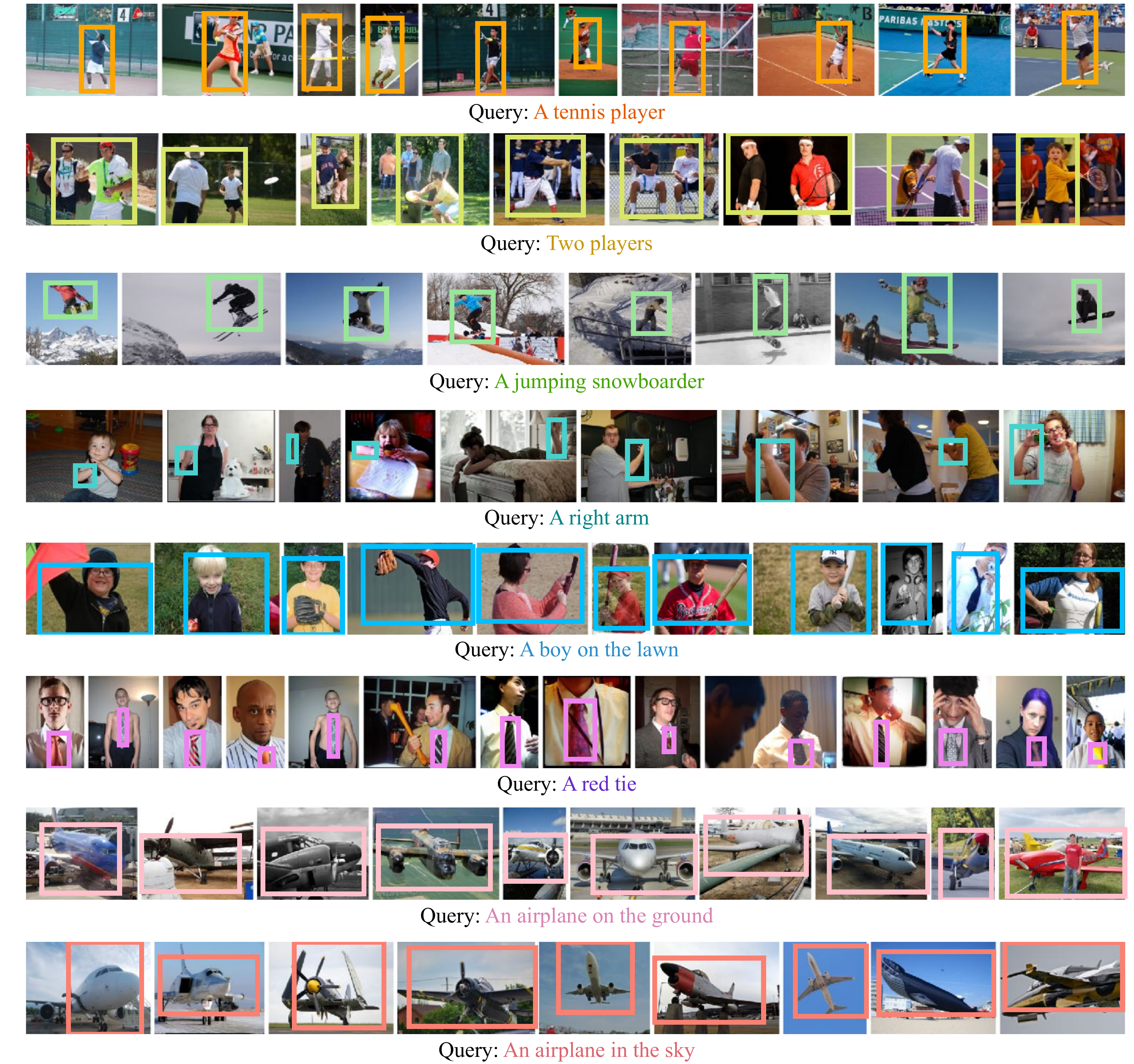}
\caption{Retrievals from COCO validation set. Top-ranked retrieval results for each query are shown.}
\label{fig:large_qual} 
\end{center} 
\end{figure*}

\end{document}